\documentclass[conference]{IEEEtran}
\IEEEoverridecommandlockouts
\usepackage{cite}
\usepackage{amsmath,amssymb,amsfonts}
\usepackage{algorithmic}
\usepackage{graphicx}
\usepackage{textcomp}
\usepackage{xcolor}
\usepackage{url}

\newcommand{\totalmodels}{21}

\def\BibTeX{{\rm B\kern-.05em{\sc i\kern-.025em b}\kern-.08em
    T\kern-.1667em\lower.7ex\hbox{E}\kern-.125emX}}

\begin{document}

\title{Cross-Model Consistency of Feature Importance in Electrospinning: Separating Robust from Model-Dependent Features}

\author{
\IEEEauthorblockN{
    1\textsuperscript{st} Mehrab Mahdian,\quad
    2\textsuperscript{nd} Ferenc Ender,\quad
    3\textsuperscript{rd} Tamas Pardy
}\\[4pt]
\IEEEauthorblockA{
    \textit{Thomas Johann Seebeck Department of Electronics},
    Tallinn University of Technology, Tallinn, Estonia \\
    mehrab.mahdian@taltech.ee
}
\quad
\IEEEauthorblockA{
    \textit{Department of Electron Devices},
    Budapest University of Technology and Economics, Budapest, Hungary \\
    ender.ferenc@vik.bme.hu
}
\quad
\IEEEauthorblockA{
    \textit{Thomas Johann Seebeck Department of Electronics},
    Tallinn University of Technology, Tallinn, Estonia \\
    tamas.pardy@taltech.ee
}
}

\maketitle
\begin{abstract}
Electrospinning is a highly sensitive process in which small variations in operating parameters can significantly affect fiber morphology. Machine learning (ML) models are increasingly used to capture these process--structure relationships and to infer feature importance. However, most studies rely on a single model, implicitly assuming that the resulting feature importance is reliable. This work evaluates the consistency of feature importance across multiple ML model families using a curated dataset of 96 polyvinyl alcohol (PVA) electrospinning experiments. Twenty-one models spanning linear, tree-based, kernel-based, neural network, and instance-based approaches were trained, and SHAP values were used to compute feature importance consistently across models. A rank-based analysis was then performed to quantify inter-model agreement. The results show that predictive performance and interpretive reliability are distinct properties. While several models achieve comparable accuracy, their feature importance rankings differ substantially. Solution concentration is identified as a fully robust parameter ($\sigma_r = 0$), whereas flow rate and applied voltage exhibit high variability ($\sigma_r > 0.9$), indicating strong model dependence. These findings demonstrate that feature importance derived from a single model may be unreliable in small datasets and highlight the need for cross-model validation in ML-based interpretation.
\end{abstract} 


\section{Introduction}

Electrospinning is a widely adopted nanofabrication technique for producing continuous
micro- and nanofibrous materials \cite{Tucker2012HistoryElectrospinning,Bhardwaj2010ElectrospinningFascinating}. Its applications span biomedical
engineering \cite{Pisani2021DesignOfExperiment}, filtration \cite{Shao2024BimodalNanofibers},
energy storage \cite{Doroudkhani2025NiOMn3O4}, and protective
textiles \cite{Wang2025MultiStructuredNanofibers}. Across these domains, application
performance is governed by the functional and mechanical properties of the nanofiber mat,
which are in turn strongly linked to structural features—including fiber diameter, diameter
distribution, morphology, and the presence of defects. These structural features are highly
sensitive to processing conditions, creating a coupled process--structure--property chain in
which small variations in parameters such as solution concentration, applied voltage, flow
rate, and tip-to-collector distance can propagate through fiber structure and ultimately
determine material performance \cite{Haghi2007Trends,Medeiros2022Modification}.
Navigating this high-dimensional, interdependent parameter space represents a central
challenge in electrospinning research \cite{Ramakrishna2005Introduction,Xue2019Electrospinning}.

Historically, optimizing electrospinning conditions has relied on iterative trial-and-error
experimentation—a slow, resource-intensive process in which researchers adjust one or more
parameters and observe the resulting fiber morphology. This approach is compounded by the
high sensitivity of the process: small changes in conditions can produce disproportionate
changes in fiber structure, making systematic manual exploration of the parameter space
impractical. Data-driven approaches offer a natural response to this challenge, promising
to model the nonlinear relationships between process inputs and structural outputs and
thereby reduce reliance on exhaustive physical experimentation.

However, the effectiveness of data-driven approaches depends critically on the quality
and structure of the available data. Electrospinning datasets are typically small,
heterogeneous, and subject to substantial variability across
studies \cite{Wang2016Environmental,Mahdian2026ElectrospinningData,Mahdian2025Controllable}—a consequence of differing experimental setups, ambient
conditions, and measurement conventions. These characteristics define a small-data regime
in which different model families may fit the same dataset equally well while learning
fundamentally different internal representations of the process--structure
relationship \cite{DAmour2020Underspecification,Breiman2001TwoCultures}. This raises a
challenge not just for predictive accuracy, but for the interpretive conclusions that
researchers draw from fitted models—conclusions that are increasingly used to guide
experimental design.

Despite these data limitations, machine learning (ML) has emerged as a powerful tool for
modeling electrospinning process--structure relationships. Recent efforts have demonstrated
high predictive performance in estimating nanofiber characteristics from experimental
inputs \cite{Khan2025PredictiveModel,CuahuizoHuitzil2023ArtificialNeuralNetworks,Roldan2026FibreCastML}, and emerging inverse design frameworks have begun to identify
process parameter combinations that yield defined target fiber
properties \cite{Mahdian2026InverseDesign,Roldan2026SpinCastML,Subeshan2026MLGA},
further reducing reliance on trial-and-error experimentation. The availability of
structured, machine-learning-ready datasets has enabled more systematic exploration of
these relationships \cite{Sarma2023ElectrospunFEAD,Mahdian2026Cogni,Mahdian2026ElectrospinningData}. Within inverse design pipelines in particular, the
surrogate model's recommendations are only as trustworthy as its internal representation
of the process--structure relationship: a surrogate that misattributes which parameters
govern fiber structure will propagate that misattribution directly into the parameter sets
it recommends. This places a premium not only on predictive accuracy but on
interpretability—on whether the feature importance a model reports reflects the true
process--structure relationship or merely the model's own inductive biases. Beyond inverse
design, there is increasing interest in using ML to extract interpretable insights about
the relative importance of process parameters—insights that researchers routinely treat as
physical evidence about which variables govern fiber structure and, ultimately, material
performance.

However, this interpretive step—critical both for scientific understanding and for the reliability of inverse design surrogates—carries an assumption that is rarely examined: that the feature importance derived from a chosen model faithfully reflects the underlying process–structure–property relationships rather than the model’s own inductive biases. To the best of our knowledge, no prior work has systematically evaluated whether feature importance conclusions are consistent across different model families in electrospinning.

In practice, different model families may assign substantially different importance to the
same parameters, and in small experimental datasets this risk is amplified. If a
researcher concludes from a single model that, say, flow rate is a dominant driver of
fiber diameter and redesigns experiments accordingly, the consequence of an unreliable
interpretation is not merely academic—it is wasted experimental effort in a process that
is already costly and time-consuming.

This raises a critical but underexplored question: \textit{can we trust the feature
importance conclusions drawn from ML models in small electrospinning datasets?} To address
this, the present work investigates the consistency of feature importance across multiple
ML model families using a curated dataset of 96 polyvinyl alcohol (PVA) electrospinning
experiments. We train diverse models spanning linear, tree-based, kernel-based, and neural
network approaches, compute SHAP (SHapley Additive exPlanations) values consistently
across all models, and quantify the agreement and variability of the resulting feature
rankings. This cross-model analysis provides a principled basis for distinguishing between
robust features—those consistently identified as important regardless of model choice—and
model-dependent features whose apparent importance reflects the choice of model rather
than the underlying process--structure--property relationships.

This study advances ML-based interpretation in electrospinning by shifting the focus from predictive performance to the reliability of model-derived feature importance. It provides empirical evidence distinguishing truly influential process parameters from those whose importance is model-dependent, particularly in small experimental datasets. These findings have direct implications for how ML models should be interpreted in electrospinning and underscore the need for cross-model validation when identifying parameters that govern fiber structure. The proposed approach is broadly applicable to other domains where ML is used to infer feature importance from limited data, offering a framework for assessing the reliability of interpretive conclusions under model uncertainty.

\section{Methods}

\subsection{Dataset}

\begin{figure}[!t]
    \centering
    \includegraphics[width=\columnwidth]{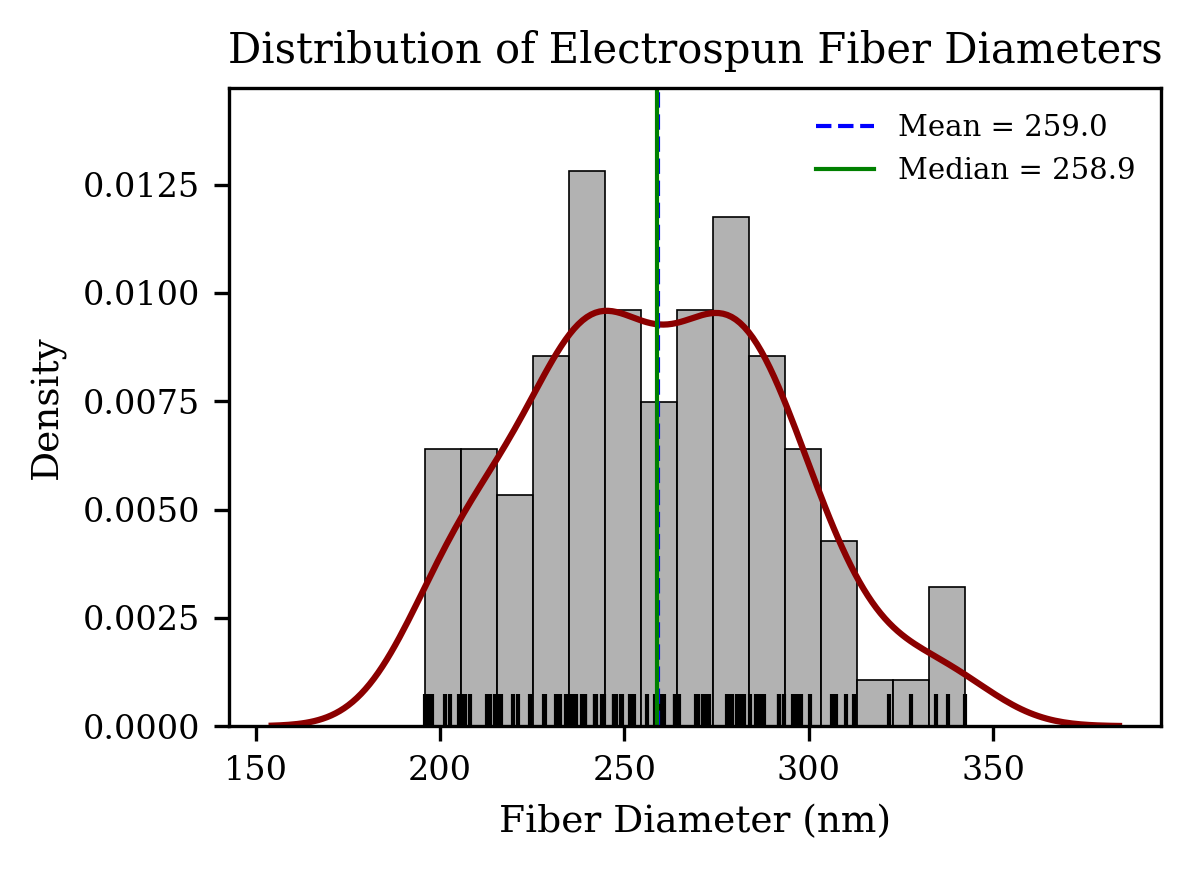}
    \caption{Measured nanofiber diameter distribution for the 96 PVA 
    electrospinning experiments (mean = 259\,nm, SD = 34.5\,nm).}
    \label{fig:diameter_dist}
\end{figure}

The dataset consists of 96 polyvinyl alcohol (PVA) electrospinning experiments 
originally reported by Ziabari et al.~\cite{Ziabari2010NewApproach} and available through 
Cogni-e-SpinDB 1.0~\cite{Mahdian2026Cogni,Mahdian2025CogniESpinDB}. It was constructed using a Response 
Surface Methodology (RSM) framework with four factors each sampled at four levels 
(Table~\ref{tab:dataset}), providing structured coverage of the process parameter 
space. The four input features—solution concentration, applied voltage, flow rate, 
and tip-to-collector distance—span the primary controllable inputs along the 
process--structure axis. The target output is mean fiber 
diameter measured from SEM micrographs, ranging approximately from 200\,nm to 
350\,nm (Fig.~\ref{fig:diameter_dist}). Ambient conditions were held constant 
throughout, making concentration a direct proxy for solution viscosity. All 
features were standardized prior to modeling. With 96 samples, this dataset is 
representative of the small-data regimes common in electrospinning research—and 
precisely the condition under which the reliability of ML-derived feature 
importance is most at risk.

\begin{table}[!t]
\caption{Experimental factors and level settings for the 96-run RSM design.}
\label{tab:dataset}
\centering
\footnotesize
\setlength{\tabcolsep}{3.5pt}
\renewcommand{\arraystretch}{1.1}
\resizebox{\columnwidth}{!}{%
\begin{tabular}{lccccc}
\hline
\textbf{Parameter} & \textbf{Unit} & \multicolumn{4}{c}{\textbf{Levels}} \\
\cline{3-6}
& & $-1$ & $-0.5$ & $0$ & $+1$ \\
\hline
Solution concentration    & wt\%  & 8    & 9    & 10   & 12   \\
Applied voltage           & kV    & 15   & 20   & 22.5 & 25   \\
Flow rate                 & mL/h  & 0.20 & 0.25 & 0.30 & 0.40 \\
Tip-to-collector distance & cm    & 10   & 12.5 & 15   & 20   \\
\hline
\end{tabular}%
}
\end{table}

\subsection{Machine Learning Models}
To investigate whether feature importance conclusions are model-dependent, models 
spanning five families were employed, covering fundamentally different inductive 
biases and learning mechanisms. \textbf{Linear models} (Linear Regression, Ridge, 
Lasso, Elastic Net) assume additive, independent parameter effects and establish 
a baseline under a linearity constraint. \textbf{Tree-based ensembles} (Random 
Forest, Gradient Boosting, Histogram-based Gradient Boosting, XGBoost, LightGBM, 
AdaBoost, Decision Tree) capture nonlinear interactions and higher-order parameter 
couplings, and are known to perform well on small structured datasets. 
\textbf{Kernel-based models} (SVR with RBF and polynomial kernels) operate in a 
transformed feature space, potentially assigning importance differently when 
parameter interactions are smooth and continuous. \textbf{Neural networks} (MLP 
with 128$\times$64 and 64$\times$32 configurations) test whether distributed, 
nonlinear representations produce consistent rankings with other families, despite 
the constraints imposed by the small dataset size. \textbf{Instance-based models} 
(KNN) assign importance based on local neighborhood structure, providing a further 
contrast to parametric approaches.

In total, \totalmodels{} models were trained. Default or minimally tuned 
hyperparameters were used throughout to ensure that observed differences in feature 
importance reflect model family characteristics rather than configuration choices. 
TreeSHAP was applied to tree-based models and KernelSHAP to all others.

\subsection{Model Evaluation}

Model performance was evaluated using 5-fold cross-validation, with mean $R^2$ 
reported across folds as the primary metric:

\begin{equation}
    R^2 = 1 - \frac{\sum_{i=1}^{n}(y_i - \hat{y}_i)^2}{\sum_{i=1}^{n}(y_i - 
    \bar{y})^2}
    \label{eq:r2}
\end{equation}

\noindent where $y_i$, $\hat{y}_i$, and $\bar{y}$ are the observed, predicted, 
and mean fiber diameters respectively. Performance evaluation serves a dual 
purpose here: it confirms that models under comparison achieve comparable 
predictive accuracy—a necessary condition for a meaningful reliability 
analysis—and it establishes that high predictive performance alone does not 
guarantee consistent feature importance, which is the central argument of 
this work.

\subsection{Explainability and Feature Importance}

SHAP (SHapley Additive exPlanations)~\cite{Lundberg2017UnifiedApproach} was used to compute 
feature importance consistently across all model families. Rooted in cooperative 
game theory, SHAP provides a theoretically grounded attribution framework that 
is model-agnostic, making it suitable for cross-model comparison. For each 
trained model, the mean absolute SHAP value was aggregated across samples to 
obtain a global importance score for each feature:

\begin{equation}
    \phi_j = \frac{1}{n} \sum_{i=1}^{n} \left| \phi_j^{(i)} \right|
    \label{eq:shap}
\end{equation}

\noindent where $\phi_j^{(i)}$ is the SHAP value of feature $j$ for sample $i$. 
Features were then ranked in descending order of $\phi_j$ for each model, 
abstracting away differences in prediction scale and enabling direct comparison 
of importance ordering across model families.

\subsection{Feature Importance Reliability Analysis}

The core of this study is a rank-based reliability analysis that quantifies 
the consistency of feature importance across all trained models. For each model, 
features were ranked by their global SHAP score $\phi_j$ (rank 1 = most 
influential). Per-model rankings were then compared using four metrics: 
\textbf{mean rank} ($\bar{r}_j$), reflecting overall tendency to be identified 
as influential; \textbf{standard deviation of rank} ($\sigma_{r_j}$), where low 
variance indicates a robust, model-independent feature and high variance indicates 
model-dependence; \textbf{Spearman rank correlation}, quantifying pairwise 
agreement between model families; and \textbf{top-$k$ agreement}, measuring the 
proportion of models that agree on the most important features. 
Together, these metrics place each feature along a reliability spectrum—from 
\textit{robust} features whose importance is consistent regardless of model 
choice, to \textit{model-dependent} features whose apparent importance reflects 
the chosen model rather than the underlying process--structure relationship.

\subsection{Implementation}

All models were implemented in Python 3.10 using \texttt{scikit-learn}, 
\texttt{xgboost}, and \texttt{lightgbm}, with SHAP values computed via the 
\texttt{shap} library (TreeSHAP for tree-based models, 
KernelSHAP for all others). The reliability analysis used \texttt{scipy} 
and \texttt{numpy}.

\section{Results}

\subsection{Model Performance Comparison}

\begin{figure}[htbp]
\centering
\includegraphics[width=\linewidth]{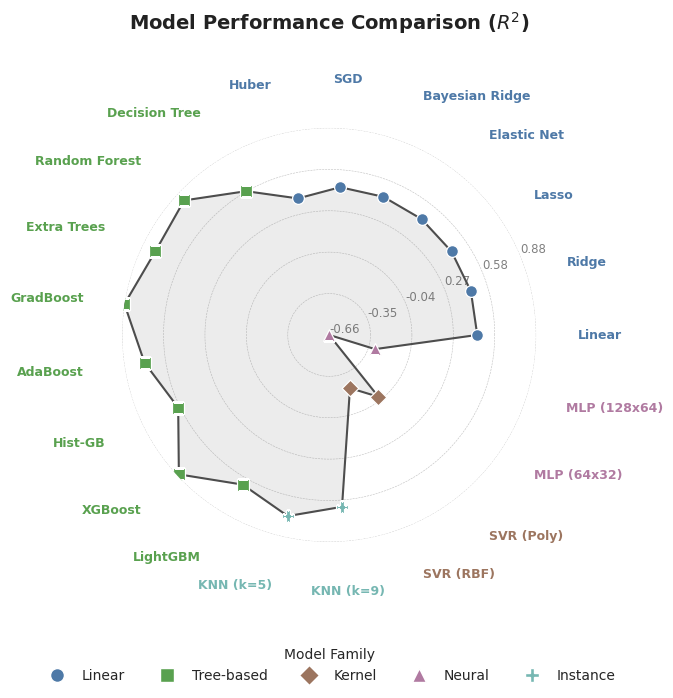}
\caption{Cross-validated $R^2$ performance of all evaluated models, grouped by 
model family. Tree-based ensembles consistently achieve the highest predictive 
accuracy, while linear models and instance-based methods show moderate 
performance. Neural network results are less stable, consistent with the 
limitations of small training sets.}
\label{fig:performance}
\end{figure}

The predictive performance of all evaluated models, assessed via 5-fold 
cross-validation, is summarized in Fig.~\ref{fig:performance}. Tree-based 
ensemble methods achieved the highest $R^2$ scores, approaching 0.9, reflecting 
their capacity to capture the nonlinear interactions inherent to the 
process--structure relationship in electrospinning. Linear models achieved 
moderate performance, while neural network approaches exhibited lower and less 
stable results across folds—consistent with the known sensitivity of 
over-parameterized models to small training sets. Kernel-based and instance-based 
methods occupied an intermediate position.

Critically, despite this performance gradient, a substantial subset of models 
from different families achieved comparable predictive accuracy. This is a 
necessary condition for the reliability analysis that follows: if models agree 
on predictions but disagree on feature importance, the disagreement must reflect 
differing internal representations of the process--structure relationship rather 
than differences in predictive quality—setting the stage for the central question 
of whether models that agree on \textit{what} to predict also agree on 
\textit{why}.

\subsection{Feature Importance Across Models}

\begin{figure}[htbp]
\centering
\includegraphics[width=\linewidth]{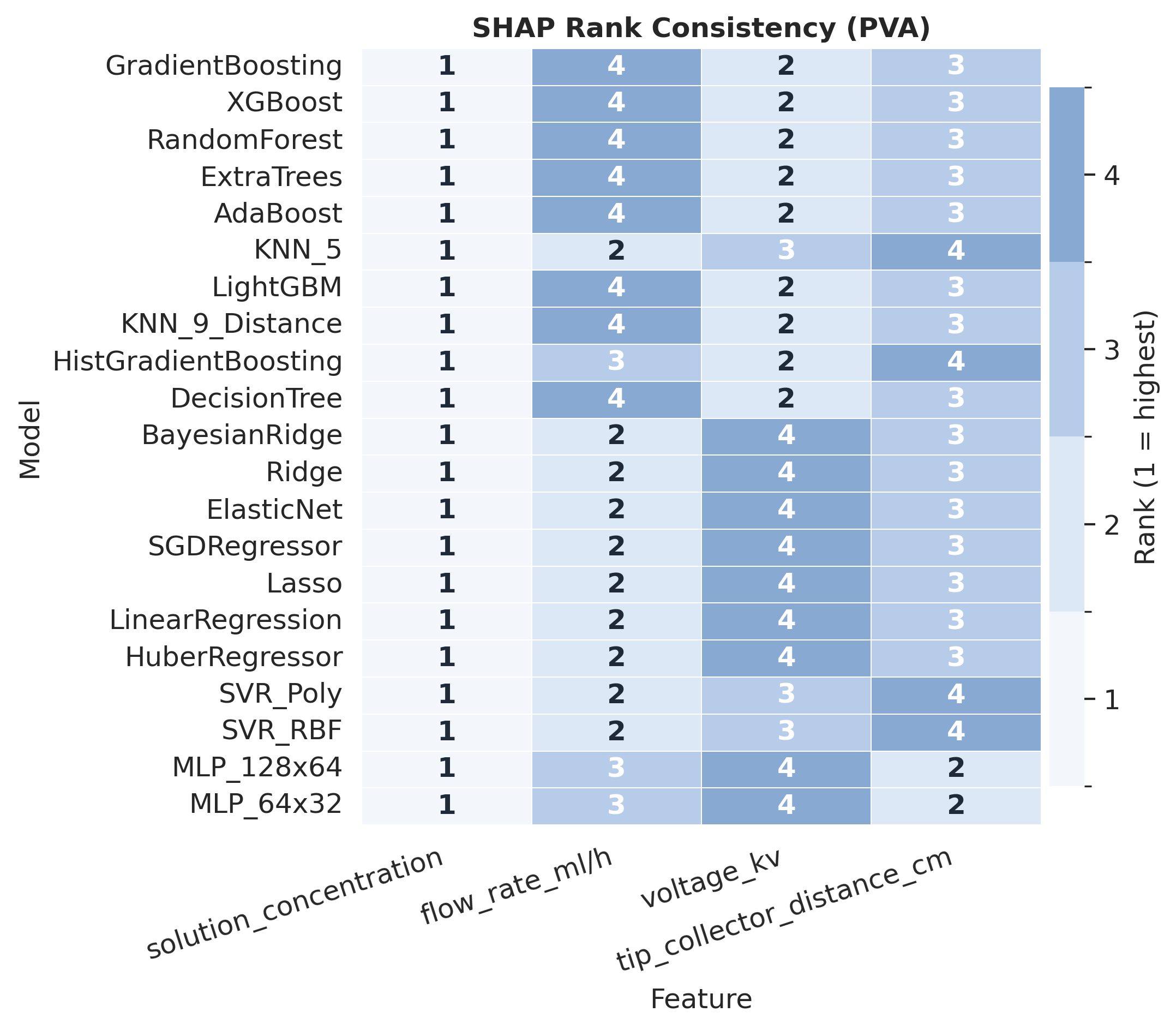}
\caption{SHAP-based feature importance rank matrix across all evaluated models. 
Each cell indicates the rank assigned to a feature by a given model (rank 1 = 
most important). Consistent coloring across a row indicates a robust feature; 
variable coloring indicates model-dependent importance.}
\label{fig:shap_rank}
\end{figure}

The SHAP-based feature importance rank matrix across all evaluated models is 
presented in Fig.~\ref{fig:shap_rank}. Solution concentration is ranked as the 
most influential feature across every model without exception—unanimous agreement 
across all five model families. This aligns with domain knowledge, as 
concentration directly controls solution viscosity and polymer chain entanglement, 
both critical determinants of jet stability and fiber 
formation.

Applied voltage and tip-to-collector distance generally appear among the more 
influential features, but with notable variation in their relative ordering 
between models—suggesting meaningful but context-dependent effects on fiber 
diameter. Flow rate presents the most variable ranking, ranging from moderately 
important to least influential depending on the model family, signaling that its 
apparent importance may reflect model choice rather than a reliable physical 
relationship.

\subsection{Feature Importance Reliability Analysis}
Table~\ref{tab:reliability} summarizes the mean rank and standard deviation for 
each feature across all models. Solution concentration achieves a mean rank of 
1.000 with zero variance—the strongest possible reliability signal—confirming 
that no model assigns any other feature higher importance. Flow rate and applied 
voltage exhibit the highest rank variance ($\sigma_r = 0.921$ and $0.926$ 
respectively), indicating that their apparent importance is sensitive to model 
choice and should not be treated as a reliable indicator of physical influence 
in a dataset of this size. Tip-to-collector distance shows lower variance 
($\sigma_r = 0.526$), placing it in an intermediate reliability category despite 
its lower mean rank.

\begin{table}[htbp]
\caption{Rank-based reliability statistics for each process parameter across all 
evaluated models (rank 1 = most important).}
\label{tab:reliability}
\centering
\begin{tabular}{lcc}
\hline
\textbf{Feature} & \textbf{Mean Rank} $\bar{r}_j$ & 
\textbf{Std. Dev.} $\sigma_{r_j}$ \\
\hline
Solution concentration    & 1.000 & 0.000 \\
Flow rate                 & 2.905 & 0.921 \\
Applied voltage           & 3.000 & 0.926 \\
Tip-to-collector distance & 3.095 & 0.526 \\
\hline
\end{tabular}
\end{table}

The mean pairwise Spearman rank correlation across all model pairs is 0.584, 
indicating moderate overall agreement in feature ordering. Higher inter-model 
correlation is observed within the tree-based family, while lower correlations 
are found between fundamentally different families—particularly linear and neural 
network models. Top-$k$ agreement analysis shows perfect consensus on the top-1 
feature (agreement = 1.000), dropping to 0.778 ($\pm$0.157) for top-3, 
confirming that reliable interpretation is strongest at the highest-importance 
position and decreases for lower-ranked features.

Based on these findings, the four parameters can be placed along a reliability 
spectrum: \textbf{solution concentration} is fully robust ($\bar{r} = 1.000$, 
$\sigma_r = 0.000$); \textbf{tip-to-collector distance} is moderately stable 
($\sigma_r = 0.526$); and \textbf{flow rate} and \textbf{applied voltage} are 
model-dependent ($\sigma_r > 0.9$), with their relative ordering varying 
substantially across model families.
\section{Discussion}

The results surface a distinction that is easy to overlook in applied machine 
learning: a model that predicts well does not necessarily interpret correctly, 
and two models that predict equally well may offer contradictory accounts of 
why. In electrospinning, where feature importance is routinely used to infer 
which process parameters govern fiber structure and to prioritize experimental 
effort, this distinction has direct practical consequences.

The unanimous identification of solution concentration as the dominant parameter 
is the study's strongest finding. Its zero-variance rank reflects a cross-model 
consensus grounded in electrospinning physics: concentration governs solution 
viscosity and polymer chain entanglement, directly controlling jet resistance to 
electrostatic stretching. The intermediate 
reliability of applied voltage and tip-to-collector distance is similarly 
interpretable—both influence fiber diameter through coupled, nonlinear mechanisms 
involving jet elongation and solvent evaporation, which different model families 
may represent with slightly different emphasis. The model-dependent behavior of 
flow rate carries the most practical consequence: a researcher relying on a 
single model might either prioritize or dismiss it, with direct implications for 
experimental design in a process that is already costly and time-consuming. 
Whether flow rate's inconsistency reflects a genuinely weaker physical role or 
an artifact of variance partitioning in small datasets remains an open question.

More broadly, this study illustrates that small-data regimes amplify the risk 
of model-dependent conclusions, and that cross-model agreement provides a more 
trustworthy basis for interpretation than single-model assertion.

\subsection{Limitations}

The dataset comprises 96 samples from a single polymer system (PVA) under 
controlled ambient conditions, limiting generalizability to other materials or 
experimental configurations. Environmental variables such as temperature and 
humidity—known to influence fiber morphology—were not included. SHAP quantifies 
associations learned by models rather than causal relationships, a distinction 
particularly relevant in RSM designs where parameter levels are not fully 
independently randomized. Finally, default hyperparameters were used to ensure 
fair comparison, which may underrepresent each model family's maximum predictive 
capability.

\subsection{Future Work}

Applying the reliability analysis to larger, multi-polymer datasets would 
establish whether the patterns observed here generalize beyond PVA. Incorporating 
environmental variables would provide a more complete characterization of the 
process--structure relationship. Methodologically, extending the approach to classification tasks—such 
as morphology prediction or instability detection—would broaden its applicability 
to other electrospinning modeling scenarios.

\section{Conclusion}

This study examined whether machine learning models agree on which 
electrospinning process parameters are most influential when applied to the same 
small experimental dataset. Solution concentration emerged as the only fully 
robust parameter, unanimously ranked as most influential across all model 
families—a conclusion grounded in its known role in governing solution viscosity 
and jet dynamics. Applied voltage and tip-to-collector distance showed moderate 
reliability, while flow rate exhibited substantial rank variability, signaling 
that conclusions about its importance should not be drawn from any single model.

The central message is that predictive performance and interpretive reliability 
are distinct properties of a machine learning model, and that the latter requires 
explicit assessment. In small experimental datasets, feature importance derived 
from a single model carries an unacknowledged risk of being model-dependent 
rather than process-driven, and hence, interpretability must be validated and not assumed. The cross-model analysis presented here offers a 
straightforward approach to quantifying this risk—applicable whenever multiple 
models achieve comparable predictive accuracy on the same dataset, in 
electrospinning and beyond.

\section*{Acknowledgment}
The authors would like to thank the Estonian Research Council for funding this research under grant PSG897.

\section*{Code and Data Availability}
The dataset used in this work is publicly available on Zenodo at \url{https://doi.org/10.5281/zenodo.16731638}; the primary data descriptor is reported in \cite{Mahdian2026Cogni}. The underlying code of this study is available at \url{https://github.com/taltechloc/Cogni-E-Spin-FIRE.git}.

\bibliographystyle{unsrt}

\vspace{12pt}

\end{document}